\documentclass[conference]{IEEEtran}
\usepackage{cite}

\ifCLASSINFOpdf
  \usepackage[pdftex]{graphicx}
\else
\fi
\usepackage{amsmath,amsthm,amscd,amssymb,latexsym}
\usepackage{url}

\usepackage{bm}
\usepackage[ruled,vlined,linesnumbered]{algorithm2e}
\usepackage{comment}

\begin{document}
%
\title{
Computer-Aided Colorectal Tumor Classification in NBI Endoscopy Using CNN Features}

\author{
Toru Tamaki, Shoji Sonoyama,\\
Tsubasa Hirakawa, Bisser Raytchev,\\
Kazufumi Kaneda, Tetsushi Koide\\
Hiroshima University\\
{\tt\small tamaki@hiroshima-u.ac.jp}
\and
Shigeto Yoshida, Hiroshi Mieno\\
Hiroshima General Hospital of West Japan Railway Company\\
Shinji Tanaka\\
Hiroshima University Hospital
}


%


\maketitle

\begin{abstract}
In this paper we report results
for recognizing colorectal NBI endoscopic images
by using features extracted from 
convolutional neural network (CNN).
In this comparative study,
we extract features from different layers from different
CNN models, and then train linear SVM classifiers.
Experimental results with 10-fold cross validations
show that features from first few convolution layers are
enough to achieve similar performance (i.e., recognition rate of 95\%) with non-CNN local features such as Bag-of-Visual words, Fisher vector, and VLAD.
\end{abstract}


%
\IEEEpeerreviewmaketitle

\section{Introduction}

Deep learning with convolutional neural network (CNN) has emerged in the last decade and rapidly made its progress in these few years, and been applied to many computer vision tasks with successful results. Different architectures of CNN have been introduced for different applications with different training strategies and datasets, however there is an interesting observation that, once trained, such CNNs can be used as discriminative feature extractor from images even for tasks that are different from original tasks where CNNs were trained.

In this paper, we investigate such features extracted from different CNNs in order to recognize colorectal endoscopic images. Prior work on endoscopic image recognition have used so-called hand-crafted features such as Bag-of-Visual Words (BoVW) \cite{Csurka2004}, Fisher vector \cite{Perronnin2010eccv,Perronnin2007CVPR,Sanchez2013ijcv}, and VLAD \cite{Jegou2010cvpr,Jegou2012pami} along with SIFT \cite{Lowe2004}, SURF \cite{Bay2007}, and wavelet \cite{Kwitt2007ICCV} or texton features \cite{Shotton2008}. Because of the variety of appearance of endoscopic images compared to texture images in standard texture datasets such as Brodatz \cite{Brodatz}, BoVW and related approaches have been successfully used for classification. We therefore compare hand-crafted features with automatic extracted features from CNN models for validating how much CNN features can be used for the endoscopic image classification task.

In section 2, we summarize relate work on NBI endoscopic image recognition as well as applications of CNN features to different recognition tasks. Section 3 explains a method examined in this paper, and experimental results are shown in section 4.

\section{Related work}

\begin{figure}
\centering
\includegraphics[width=\linewidth]{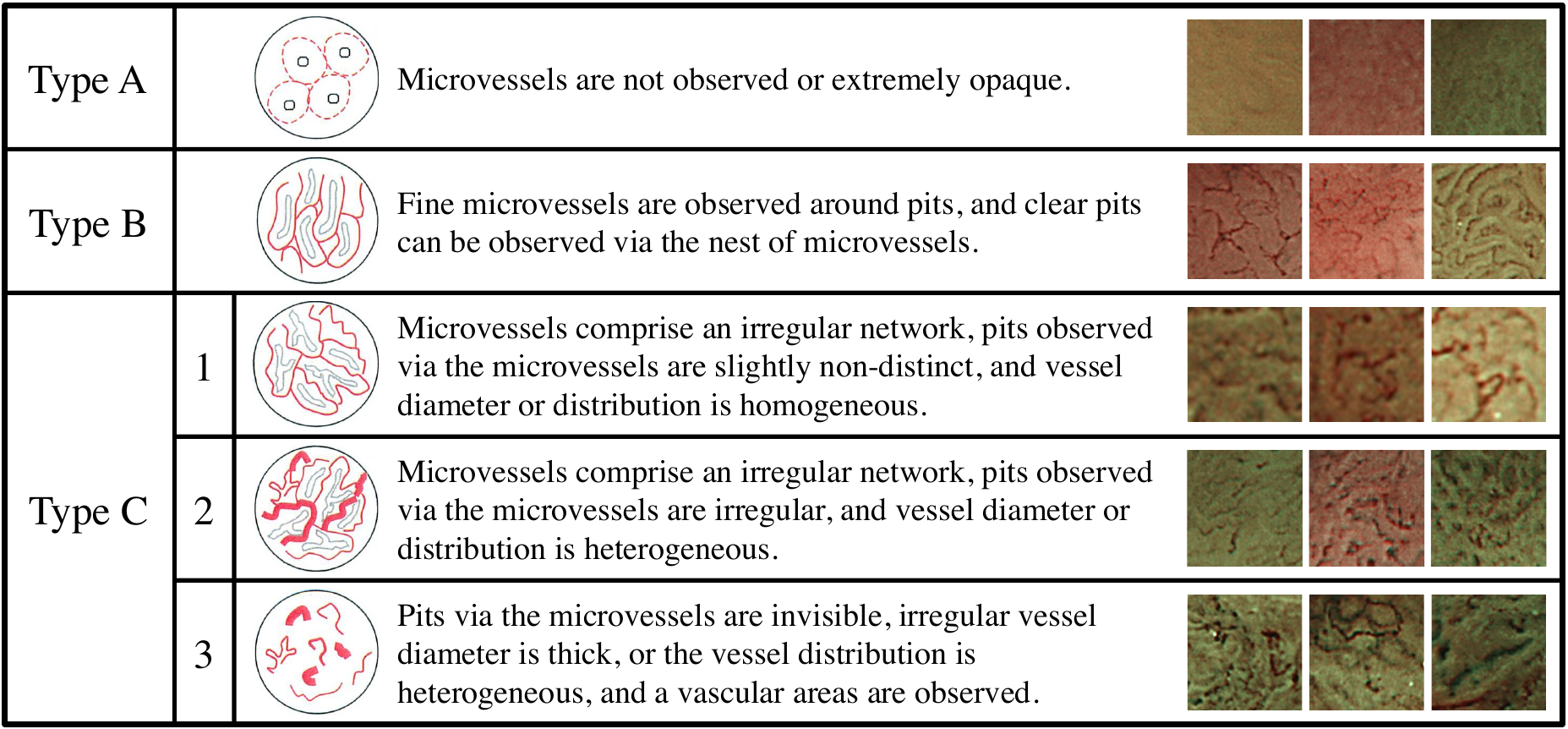}
\caption{
NBI endoscopic images and medical categories \cite{Kanao2008}.
}
\label{fig:nbi}
\end{figure}

Texture classification of colorectal cancer images taken by endoscopic examinations has been studied over years for medical diagnosis in medicine, and recently for computer-aided diagnosis in computer science.
Nowadays many hospitals perform endoscopic examinations 
with Narrow-band imaging (NBI) system which enhance vessel texture patterns
on mucosal surfaces by illuminating narrow band lights.
Before the emergence of the NBI systems, classifications for pit patterns of colorectal polyps have been developed for medical diagnosis, and similar criteria have also been developed for NBI images (see Figure \ref{fig:nbi}). These classification criteria, called NBI (magnification) findings, is however relatively difficult to apply in real situations, in particular, for non-experts and medical students on training, therefore computer-aided diagnose support systems are expected to develop.
Among many attempts, Tamaki et al. \cite{DBLP:journals/mia/TamakiYKRKYTOMT13} used the BoVW approach with densely sampled SIFT features and SVM classifiers for classifying 908 NBI images into three categories (A, B, and C3) of the NBI magnification findings, and obtained recognition results of between 94\% to 96\% accuracy (depending on SVM kernel types and scales of SIFT). Later Sonoyama et al. \cite{doi:10.1117/12.2081928} applied Fisher vector and VLAD, which are more advanced schemes for encoding local features, along with the attempt for decreasing computation cost for obtaining visual words while retaining accuracy.

Deep learning is one of neural network models with many layers stacking one another, including convolutional neural network (CNN) models where layers are convolutions of their previous layers, typically followed by fully connected layers. Because of its outperforming performances for a diverse of recognition tasks, CNN is rapidly taking its place since 2012, necessarily replacing existing schemes (e.g., BoVW) as well.
Fully training a CNN requires a large dataset as well as computation resources,
but for tasks with a smaller dataset and limited computation resources,
a fine-tuning from a pre-trained network is shown to be successful.
Moreover, features from CNN layers without any fine-tuning can be used for
different tasks with a good performance \cite{Razavian_2014_CVPR_Workshops}, and this strategy has been tested for medical image recognition tasks \cite{Bar2015,Shie2015}.

Then it is natural to test such features for the NBI image recognition task.
Thanks to the publicly available pre-trained CNN models,
we use Caffe \cite{jia2014caffe} to extract CNN features 
Our contribution is to show the comparison of performances between CNN features
from different layers from different CNN models, in contrast to existing work \cite{Razavian_2014_CVPR_Workshops,Bar2015,Shie2015} that uses only few layer features.
Discussions based on obtained results is another contribution of this paper
in order to design and propose a better CNN model for this kind of specific task: NBI colorectal polyp image recognition.

\begin{figure}
\centering
\includegraphics[width=\linewidth]{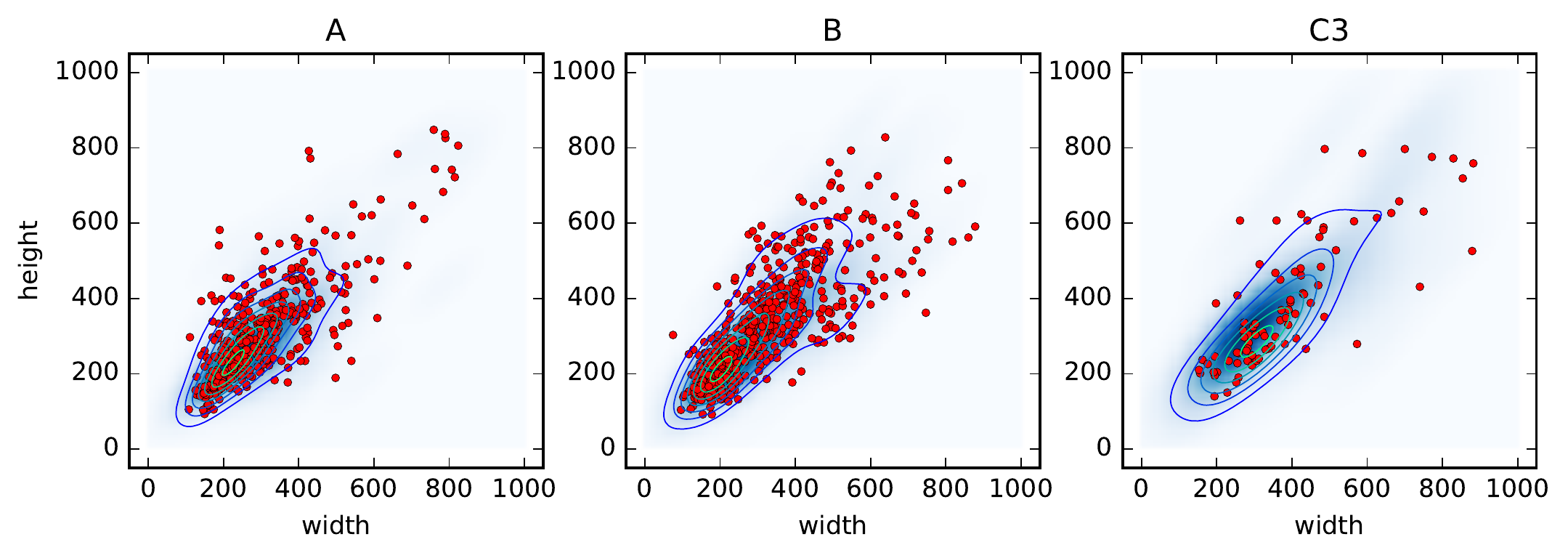}
\caption{
Size distribution of image patches for each category of NBI.
}
\label{fig:size}
\end{figure}

\begin{figure}
\centering
\includegraphics[height=20cm]{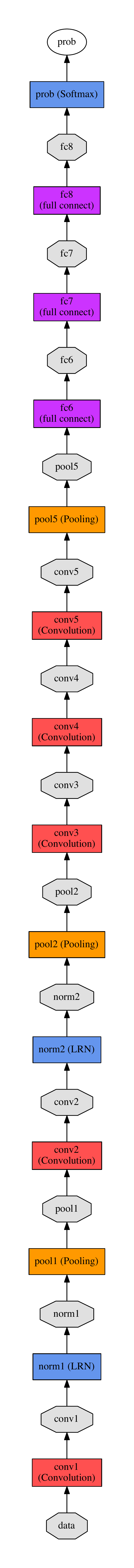}
\includegraphics[height=20cm]{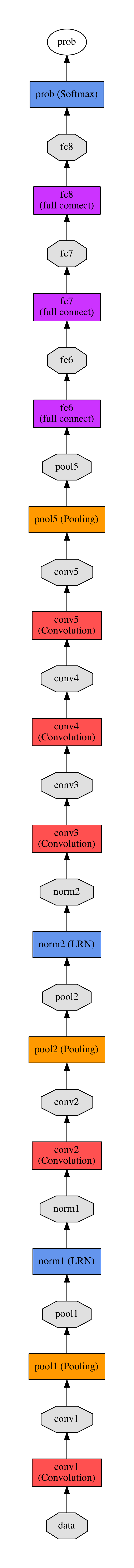}
\includegraphics[height=20cm]{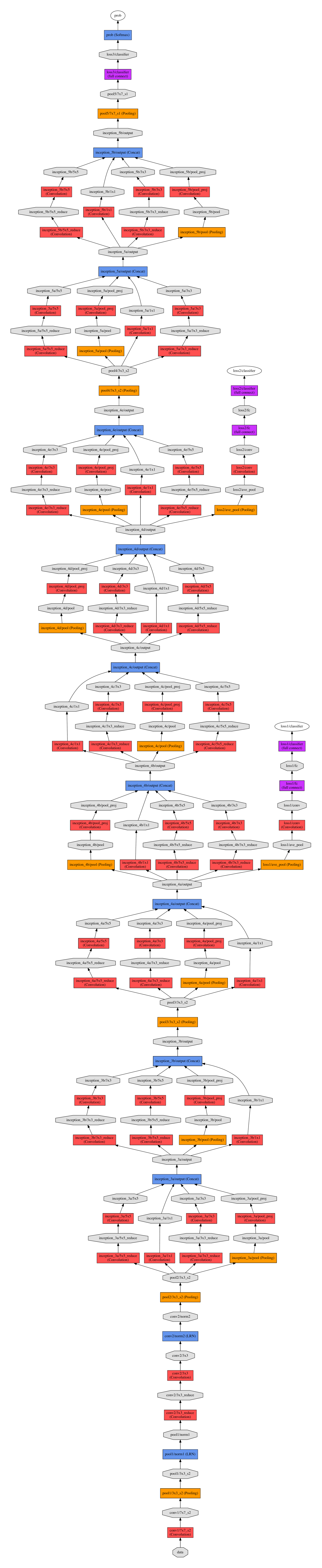}
\caption{CNNs used for experiments:
(left) AlexNet, (middle) Caffe Net, and (right) GoogLeNet.
Gray polygons (and white ellipses) represent data, or CNN features,
and color rectangles represent data processing in each layer.
Note that numbers of layers are counted differently in different literature.
Here a layer is data after a processing (such as convolution, pooling, normalization, etc.). Numbers in names of layers show a pack of these layers (for example, conv5 and pool5 are 5th layer of the CNN model in the literature).
}
\label{fig:cnn-models}
\end{figure}

\section{Method}

The target dataset for the comparison in this paper is the NBI image dataset
including 908 NBI patches collected from endoscopic examinations at Hiroshima University\footnote{The study was conducted with an approval from the Hiroshima University Hospital ethics committee, and an informed consent was obtained from the patients and/or family members for the endoscopic examination.},
and labeled based on the NBI magnification findings \cite{Kanao2008,kanao2009narrow,Oba2010}
which categorizes appearances of tumors into type A, B, and C, and type C is further sub-classified into C1, C2, and C3 based on microvessel structures (see Figure \ref{fig:nbi}). In this work we use only types A, B, and C3 by following previous work \cite{DBLP:journals/mia/TamakiYKRKYTOMT13,doi:10.1117/12.2081928,Hafner2015b,Sonoyama2015}.
A patch is trimmed from a larger frame of the entire endoscopic image, so that the trimmed rectangle region well represents the typical texture pattern of the colorectal polyp appearing in the frame.
Different patches has therefore different sizes
ranging between about 150 to 600 pixels. Distributions of sizes
for each of three NBI categories are shown in Figure \ref{fig:size}.

There might be different ways to encode these image patches with CNN features.
For example, Krizhevsky et al. \cite{NIPS2012_4824} crop an image into several patches of fixed size for data augmentation, then take the average of predictions of those patches. Razavian et al. \cite{Razavian_2014_CVPR_Workshops} employ a similar way but more aggressively in order to obtain more than one hundred patches from a single test image. Another choice might be extracting CNN features from patches provided by region proposal methods, followed by an old-style (i.e., BoVW or Fisher vector) encoding scheme \cite{cimpoi15deep}.

As a baseline, we use the simplest way: just resizing an image patch to a fixed size, and feeding it to CNN models for training and testing. This is a typical way in the literature of CNN models, because the resizing might not change the global layout of an image, which is an important cue for many tasks such as object and scene recognition. However it might not the case for NBI images because intuitively changing the scale and aspect ratio of texture images could affect the appearance of the texture pattern, hence leading to a deterioration of classification performance. In the experiments, however we will show counter-intuitive results, that is, this simple resizing works well.
After the simple resizing, we subtract from each image the average rgb pixel value
that is used when a CNN model is trained.

For comparison, we extract different CNN features from different layers including full connection layers and convolution layers before and after pooling and normalizations.
There are several ways for standardization of features (such as 0 mean and one sigma, max/min to $\pm 1$, etc.), however we don't perform any standardization and just use raw CNN features extracted to SVM classifiers with the linear kernel.

We evaluate results with 10-fold cross validation.
In each fold, the value of SVM parameter $C$ is determined by 3-fold cross validation of training samples.
We report the average recognition accuracy with standard deviation of 10 folds.

\begin{figure}[t]
\centering
\includegraphics[width=\linewidth]{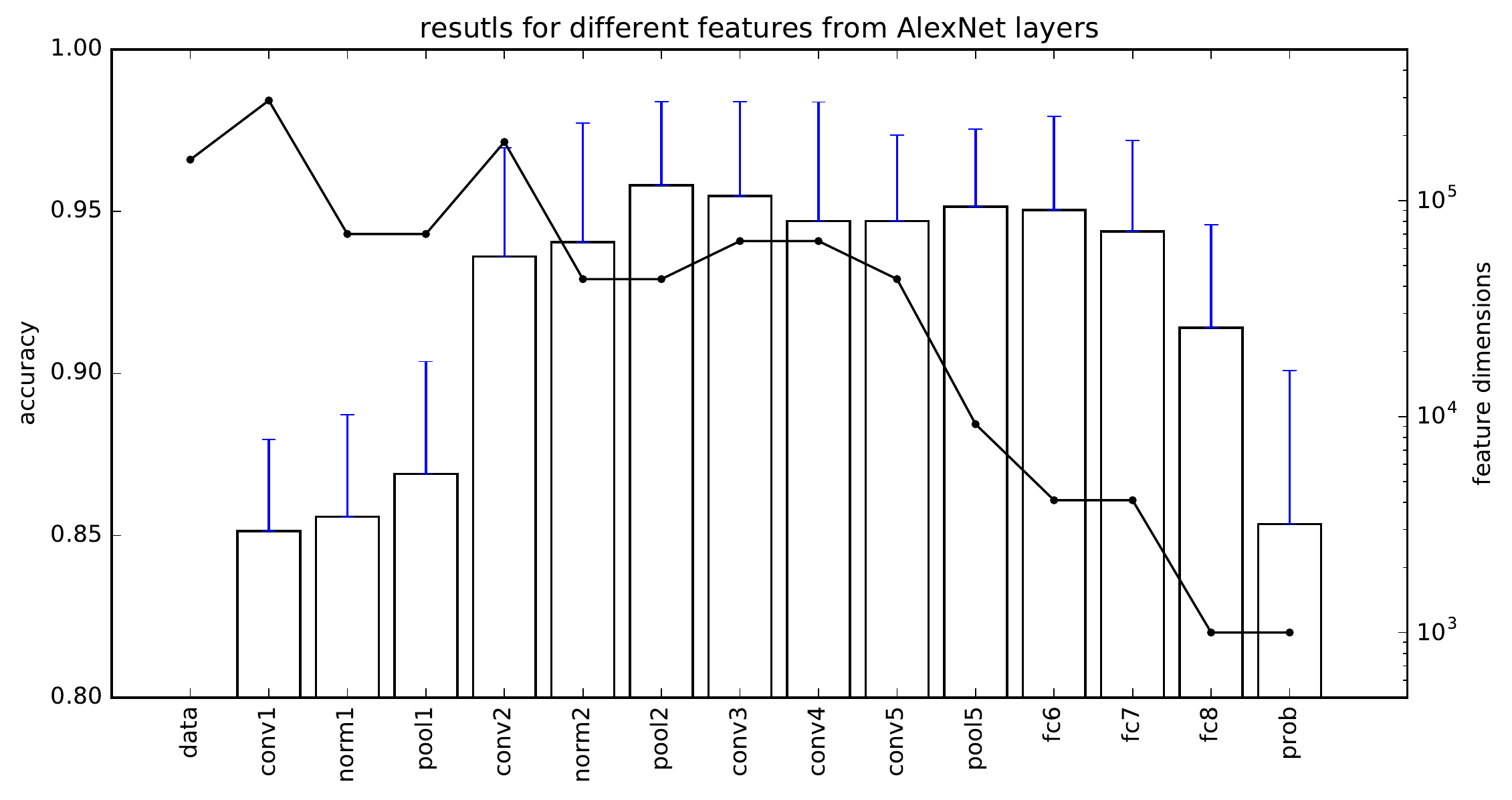}
\caption{Recognition results with AlexNet layers.
Each box represents the average accuracy (left vertical axis) for each feature,
and error bar stands for the standard deviation.
Dimension of each feature (right vertical axis)
is also shown with the black solid line.
}
\label{fig:result-alexnet}
\end{figure}

\begin{figure}[t]
\centering
\includegraphics[width=\linewidth]{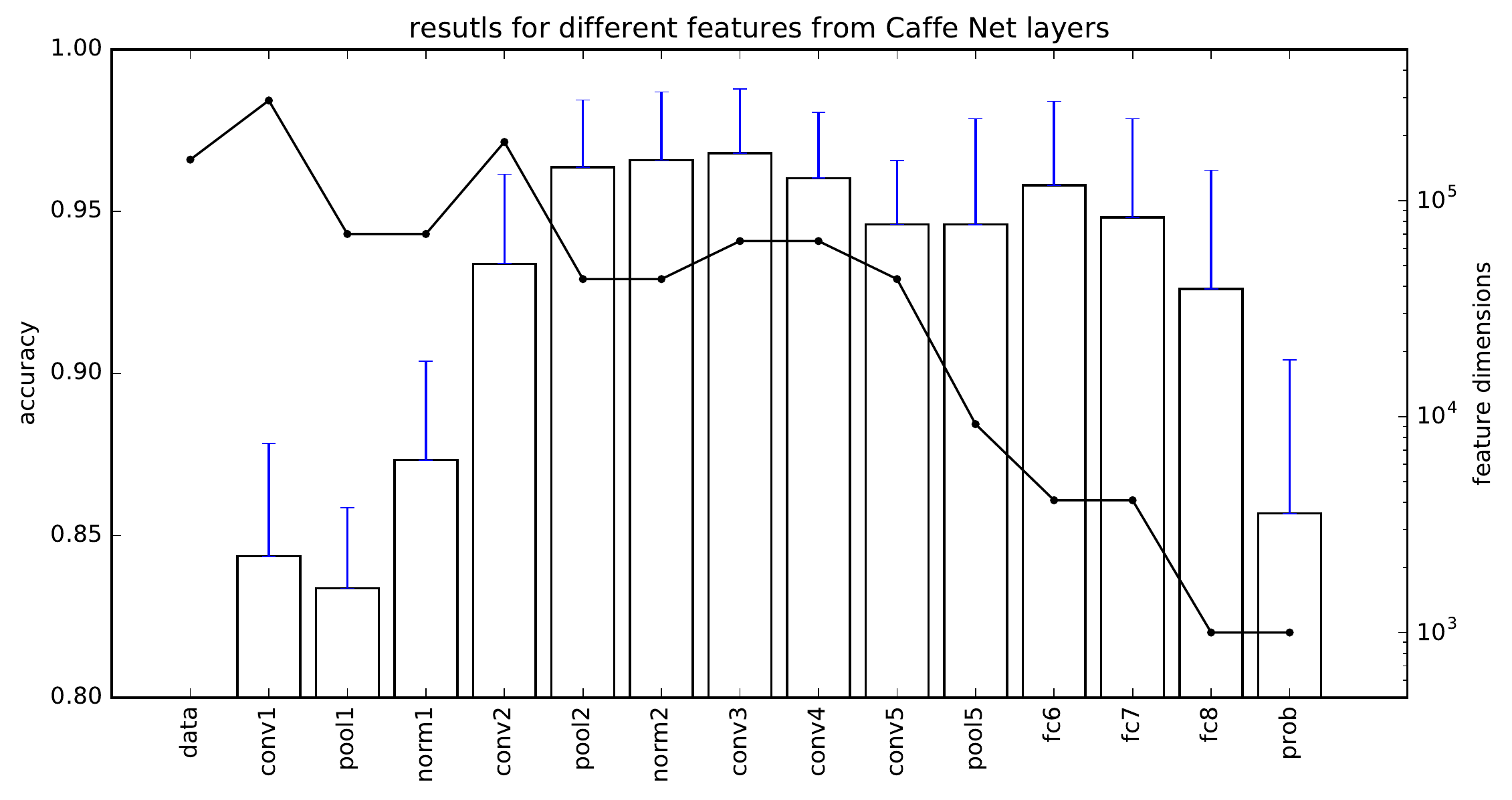}
\caption{Recognition results with Caffe Net layers.}
\label{fig:result-caffenet}
\end{figure}

\begin{figure}[t]
\centering
\includegraphics[width=\linewidth]{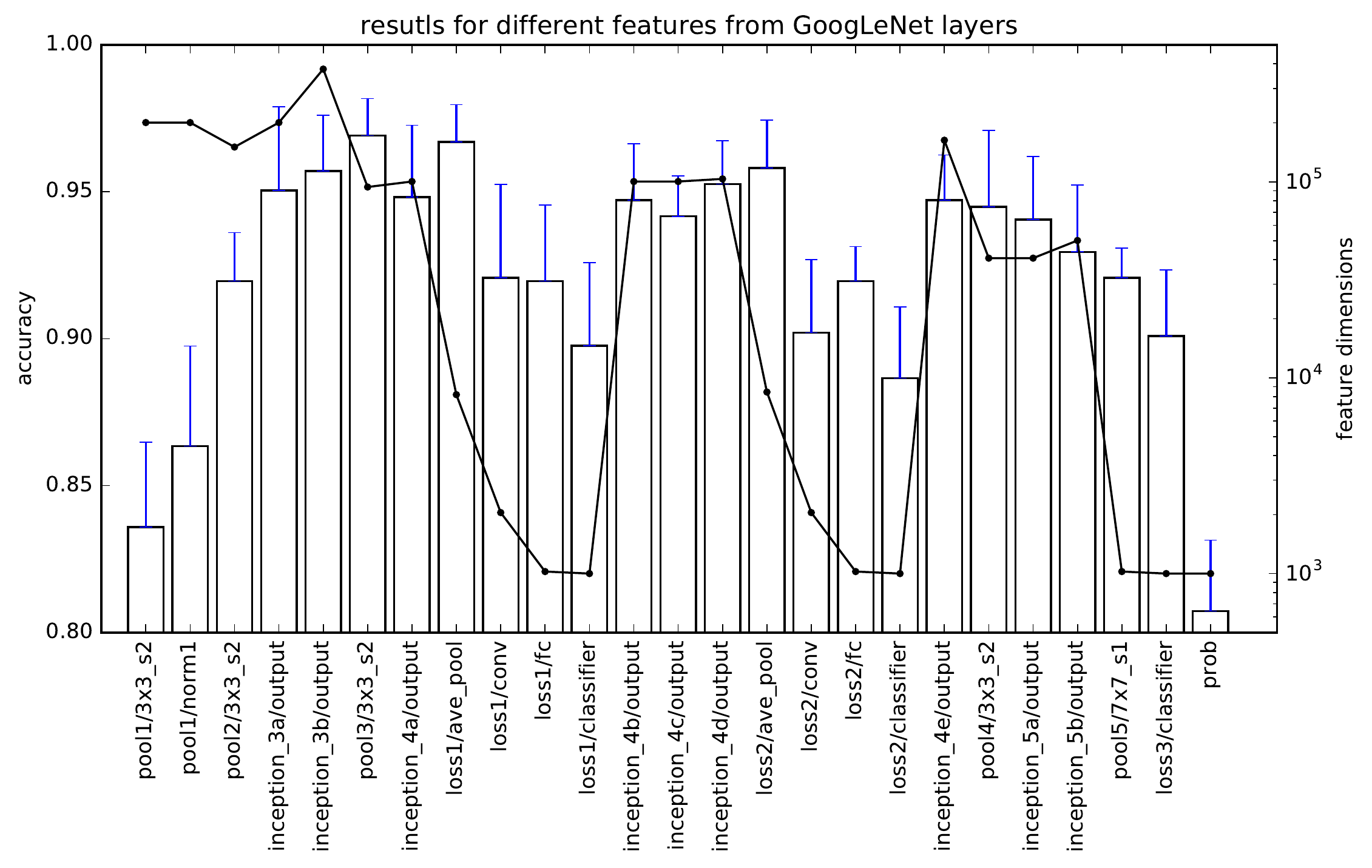}
\caption{Recognition results with GoogLeNet layers.}

\label{fig:result-googlenet}
\end{figure}

\begin{figure}[t]
\centering
\includegraphics[width=\linewidth]{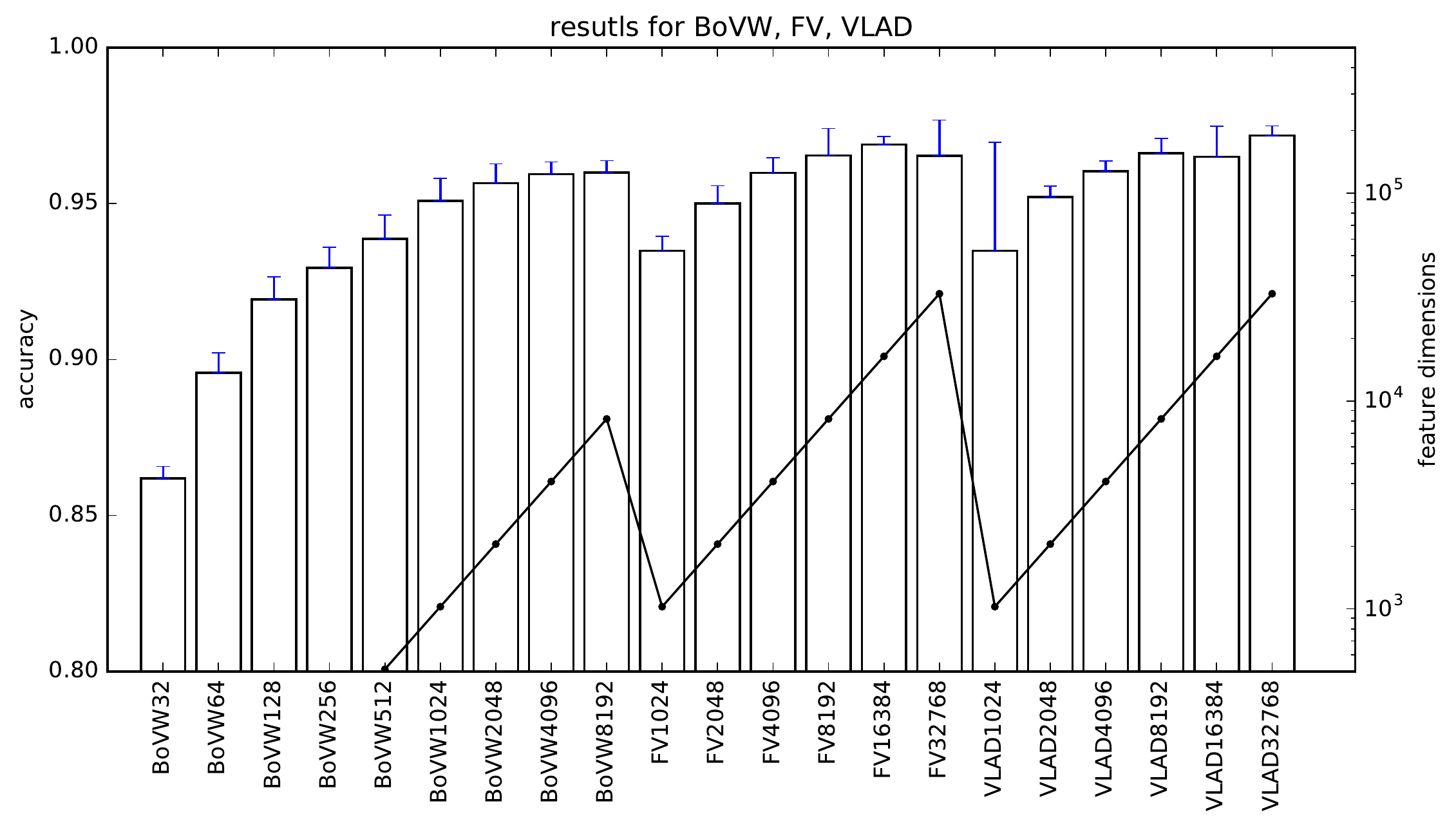}
\caption{Recognition results with BoVW, Fisher Vector (FV), and VLAD.}

\label{fig:result-vq}
\end{figure}

\section{Results}

We describe results with three different CNN models: AlexNet, Caffe Net, and GoogLeNet.
Layer layouts of these models are shown in Figure \ref{fig:cnn-models}.

\subsection{AlexNet}

Figure \ref{fig:result-alexnet} shows results for different layers of 
a replication%
\footnote{This can be obtained from
\url{http://dl.caffe.berkeleyvision.org/bvlc_alexnet.caffemodel},
and the difference from the original model is described in
\url{https://github.com/BVLC/caffe/blob/master/models/bvlc_alexnet/readme.md}.
}
of a CNN model proposed by \cite{NIPS2012_4824}, so-called AlexNet.
This has 8 layers, 5 of which are convolution layers (conv1 to conv5)
with normalization (norm1 and norm2) and pooling (pool1, pool2, and pool5),
and 2 are fully connected layers (fc6 and fc7),
followed by a softmax layer (fc8).

The maximum accuracy of 94.1 $\pm$ 3.7\%  was achieved
at pool2, the pooling after the second convolution layer,
while differences to successive layers are relatively small
compared to larger standard deviations.
In terms of dimensionality of features, CNN features of convolutional layers
are very large compared to those of fully connected layers.
The use of fc6, the first fully connected layer with smaller dimension,
is a reasonable choice to use, as done in \cite{Razavian_2014_CVPR_Workshops}.

\subsection{Caffe Net}

Figure \ref{fig:result-caffenet} shows results for different layers of Caffe Net \cite{jia2014caffe}. 
This is also a replication%
\footnote{This can be obtained from
\url{http://dl.caffe.berkeleyvision.org/bvlc_reference_caffenet.caffemodel},
and the details are described in
\url{https://github.com/BVLC/caffe/blob/master/models/bvlc_reference_caffenet/readme.md}.
}
of the AlexNet \cite{NIPS2012_4824},
having the same 8 layers, but normalization and pooling are just switched.

We can see the similar trend that the maximum performance was obtained
in the first few layers (precisely, 96.8 $\pm$ 2.0\% at conv3)
while there are no large differences between following layers.

\subsection{GoogLeNet}

Figure \ref{fig:result-googlenet} shows results for different layers of
a replication%
\footnote{This can be obtained from
\url{http://dl.caffe.berkeleyvision.org/bvlc_googlenet.caffemodel},
and the details are described in
\url{https://github.com/BVLC/caffe/blob/master/models/bvlc_googlenet/readme.md}.
}
of a CNN model named GoogLeNet \cite{Szegedy_2015_CVPR}.
This has 22 layers in which, after some early convolution layers,
the Inception modules are stacked one another while involving some pooling layers.
In addition to the final fully connected layer,
two fully connected layers are inserted in the middle of the network
for training, although these are discarded for prediction.
In this experiment, we use these two additional fully connected layers as well for comparison.

The maximum performance of 96.9 $\pm$ 1.3\% was obtained at pool3/3x3\_s2 layer,
which is the pooling layer after first two Inception modules following early convolution layers. Similar performance was obtained at loss1/ave\_pool layer after third Inception module (96.7 $\pm$ 1.3\%).

\subsection{Hand-crafted features}

Figure \ref{fig:result-vq} shows results
for three hand-crafted non-CNN based features: Bag-of-Visual Words (BoVW), Fisher vector, and VLAD.
Feature dimensions of features depend on how many visual words are used for encoding,
and how features are encoded with visual words.
Performances increase as many visual words are used, in fact, the maximum performance 
of 97.2 $\pm$ 3.2\% was obtained with VLAD of 32768 dimension.

Note that error bars in Figure \ref{fig:result-vq} indicate standard deviations 
however which is different with other figures.
Because visual words obtained from a training set are different in different fold in the 10-fold cross validation, each fold is not assumed to be equal. Therefore, we performed 10 trials of the 10-fold cross validation, and we report the average and standard deviation of the 10 trials in Figure \ref{fig:result-vq}, whereas results of a single trial of the 10-fold cross validation are shown in Figures \ref{fig:result-alexnet}, \ref{fig:result-caffenet}, and \ref{fig:result-googlenet}.

\section{Discussions}

We can see some observations in the experimental results.

First, CNN features extracted from each layer works as well as existing reports \cite{Razavian_2014_CVPR_Workshops,Bar2015,Shie2015}. In Figure \ref{fig:result-alexnet} and \ref{fig:result-caffenet} the recognition result with the data layer is shown in the left-most bar, although it is far below the visualization range (59.0 $\pm$ 4.3\%), which is quite reasonable and expected because raw intensity values are usually not expected to work for recognition. Even after one convolution layer, performances increase to at least 80\%.

Interestingly, even probability outputs (prob layers) work better then the data layer. As three CNN models used in the experiments were trained on ILSVRC 2012 dataset \cite{ILSVRC15}, the prob layer output vector of 1000 dimension is the recognition result for 1000 object categories and it is completely meaningless in this kind of experiments (hence it was excluded in \cite{Bar2015,Shie2015}). Nevertheless, the experiment shows that the output is still useful as a feature for other classification task.

Second, maximum performances were obtained at the first few convolutional layers, not the last fully connected layers. In the results of the AlexNet, pool2 layer achieves the best, and following layers (to fc7) works similarly. In Caffe Net, the performance peaks at conv3 layer then slightly decrease in the following convolutional layers, while the fully connected layer (fc6) boosts but a little.
We can see the similar tendency in the results of GoogLeNet; after the peak of pool3/3x3\_s2 layer, performances of inception modules decrease.

A good performance in early convolutional layers of CNN models can support the results with hand-crafted features in Figure \ref{fig:result-vq}. BoVW and related schemes are known to have similar structure of two convolutional layers (filtering, pooling, and nonlinear transformation). Therefore these results might suggests that a CNN with two or three convolutional layers is enough for this task of NBI image classification, and that different filter outputs are essentially important like as conventional texture analysis with wavelet and Gabor filters. Adding more convolutional layers however can blur out small texture appearances relatively small in an image patch, as performances decrease in later layers. Fully connected layers is not necessary in this sense, but has a role for dimensionality reduction because similar performances were achieved with more smaller dimensionality.

Third, pooling layers in the early stage seem to work well, as we can see pool2 in AlexNet and Caffe Net, and pool2/3x3\_s2, pool3/3x3\_s2, loos1/ave\_pool and loos2/ave\_pool in GoogLeNet. Later stages doesn't look to gain from pooling layers such as 
pool5 in AlexNet and Caffe Net, pool4/3x3\_s2 and pool5/7x7\_s1 in GoogLeNet.
Pooling layers are usually said to provide the invariance to translation for objects in the scene, however it is questionable for texture images (NBI image patches) because small translation errors obviously do not change the nature of texture appearance. Instead, we may be able to think the pooling layers are feature extractor like as more sophisticated pooling schemes \cite{Lin2013}.

Last but not least, simply resized NBI image patches are well classified with CNN features. Although changing the scale and aspect ratio of images is harmful to classification intuitively, CNN features look not being affected by those adversarial resizing. One possibility might be that patches of different category have different size distributions so that classifiers use such a size cue instead of texture cues. It is however not the case because distributions of patch sizes are similar to each other as shown Figure \ref{fig:size}. Exploring more sophisticated methods for encoding CNN features, rather than resizing, might not be necessary while further comparisons should be conducted.

\section{Conclusions}

We have reported the comparison of performances for the NBI endoscopic image classification task by using different CNN layer features from three different CNN models. Experimental results show that few convolutional layers work as a good CNN feature extractor while many layers can deteriorates the performance. We didn't do any fine-tuning, but instead put just these CNN features into linear SVM classifiers.
This lead us to come up with a CNN model with the following configuration; there are two or three convolutional layers (with fewer filters compared to AlexNet and Caffe Net (96 filters) or GoogLeNet (64 filters) in the first layer), followed by a fully connected layers which is aimed mainly to dimensionality reduction. We plan to construct such a CNN model and compare results in future work.

\bibliographystyle{IEEEtran}
\bibliography{report,references_bibdesk}

\end{document}